\documentclass[letterpaper]{article}
\usepackage[submission,draft]{aaai2027}
\usepackage[hyphens]{url}
\usepackage{graphicx}
\usepackage{natbib}
\usepackage{caption}
\makeatletter
\def\showauthors@on{T}
\makeatother

\usepackage{amsmath}
\usepackage{amssymb}
\usepackage{amsthm}
\usepackage{booktabs}
\usepackage{multirow}
\usepackage{datetime2}
\RequirePackage{fancyhdr}

\title{MATCH: Model-Aware Tool Learning with Curriculum Scheduling and Hierarchically Gated Rewards}

\author{
Shihao Liu$^{1,3}$\thanks{Work done during internship at Honor Device Co., Ltd},
Hao Yin$^{2,3*}$,
Lijun Liu$^{3}$\thanks{Corresponding authors.},
Zhengzong Chen$^{3\dagger}$,
Yuanyuan Zhao$^{3}$,
Fei Huang$^{3}$\\
$^{1}$\textnormal{School of Cyber Security, University of the Chinese Academy of Sciences}\\
$^{2}$\textnormal{University of the Chinese Academy of Sciences}\\
$^{3}$\textnormal{Honor Device Co., Ltd}\\
}
\affiliations{
    Correspondence: \{liulijun3, chenzhengzong\}@honor.com
}
\makeatletter

\def\twodigits#1{\ifnum#1<10 0\fi\the#1}

\fancypagestyle{firstpage}{
  \fancyhf{}
  \fancyfoot[C]{\thepage}
  \lhead{\includegraphics[height=20pt]{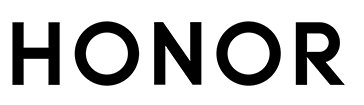}}
  \rhead{\textcolor[HTML]{4D4D4D}{\DTMtoday}}
  \renewcommand{\headrulewidth}{1pt}
  \renewcommand{\headrule}{%
    {\color[HTML]{4D4D4D}%
    \hrule\@height\headrulewidth\@width\headwidth\vskip-\headrulewidth}%
  }
  \setlength{\headheight}{1pt}   
  \setlength{\headsep}{-5pt}       
}
\makeatother
\renewcommand{\headrulewidth}{1pt}
\begin{document}
\thispagestyle{firstpage}
\maketitle
\pagestyle{fancy}

\begin{abstract}
Tool learning enables large language models (LLMs) to use external tools for tasks beyond parametric knowledge. Reinforcement learning can optimize tool-call behavior from feedback, but current methods still face two problems: fixed-threshold curricula can become misaligned with the policy's evolving capability boundary, and additive rewards can leak argument-level credit when the predicted tool is wrong.
To address these problems, we propose \textbf{MATCH}, a closed-loop framework for model-aware tool learning with curriculum scheduling and hierarchically gated rewards. \textbf{Model-Aware Curriculum Learning} (MACL) maintains reward-derived sample difficulty that co-evolves with the policy, and each epoch selects samples near the current capability boundary together with a top-$k$ pool of harder cases. \textbf{Hierarchical Tool-call Gated Reward} (HTGR) scores tool name, argument key, and argument value as a gated chain, granting credit at each level only when prerequisites hold. The same HTGR rewards drive both GRPO updates and MACL's difficulty refresh, closing the loop between policy optimization and sample scheduling.
On API-Bank and BFCL V3, MATCH reaches 72.19\% and 62.87\% overall accuracy, outperforming the main supervised and RL-based baselines. Backbone experiments further show consistent improvements across four backbones from two model families.
\end{abstract}

\section{Introduction}

\begin{figure}[!t]
    \centering
    \includegraphics[width=\columnwidth]{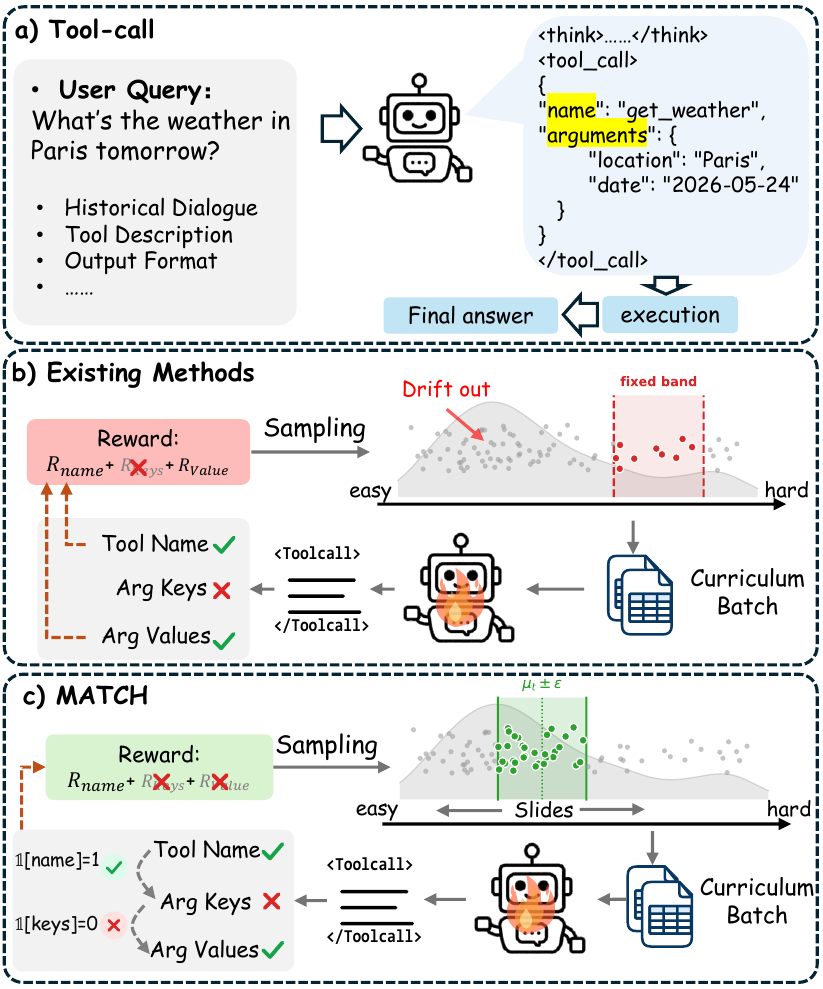}
    \caption{Motivation for MATCH. \textbf{(a)} A typical
Tool-call rollout. \textbf{(b)} Existing
RL-based tool learning, with additive reward (left) and fixed curriculum
band (right). \textbf{(c)} MATCH, with HTGR
(left) and MACL adaptive band (right).}
    \label{fig:introduction}
\end{figure}

Large language models (LLMs) have made substantial progress in reasoning, yet many real-world tasks require information and operations beyond their parametric knowledge \citep{yao2022react,schick2023toolformer,qin2024toollearningfoundationmodels}. Tool learning addresses this gap by teaching LLMs to invoke external tools during problem solving (Figure~\ref{fig:introduction}(a)). Early methods rely on prompting or supervised fine-tuning over offline trajectories \citep{chen2023afireact,qin2024toolllm,zhang2024xlam}, which imitate observed behavior but do not directly optimize tool-call outcomes. Reinforcement learning instead optimizes the policy directly from execution feedback \citep{jin2025searchr1trainingllmsreason,li2025torlscalingtoolintegratedrl,zhang2025tooln1}.

Early work on RL-based tool learning centers on reward design for generated tool calls. Outcome-level rewards provide only sparse success signals \citep{zhang2025tooln1}, while fine-grained rewards decompose correctness along tool-name, argument-key, and argument-value dimensions to give denser feedback \citep{qian2025toolrlrewardtoollearning,zeng2025toolzero}. More recent work extends reward signals into sample scheduling, with ToolSample turning reward statistics into a dynamic curriculum that prioritizes informative samples \citep{feng2025toolsample}. However, current methods still face two problems. Existing rewards ignore the dependency structure of tool execution, and existing curricula rely on fixed thresholds that do not adapt to the policy.

\begin{figure}[t]
    \centering
    \includegraphics[width=\columnwidth]{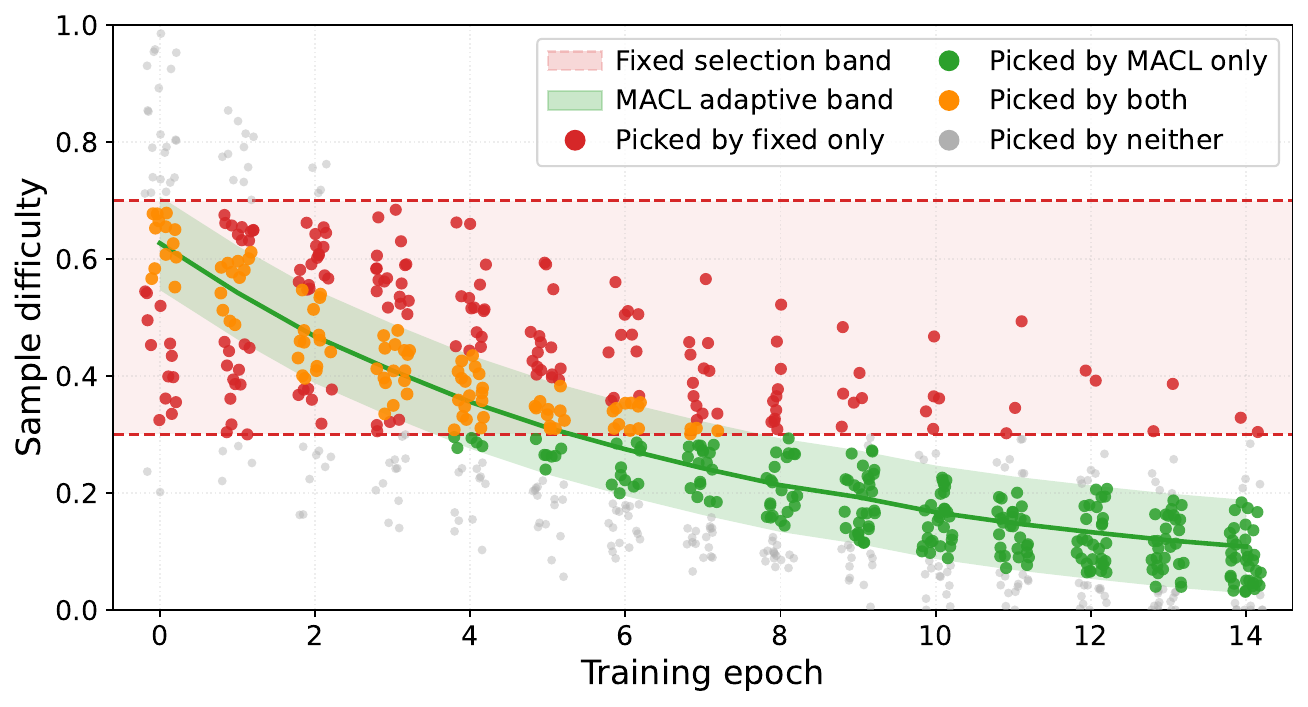}
    \caption{Sample-difficulty dynamics on a fixed dataset. The bulk descends as the policy improves. A fixed band (pink) drifts off the informative region after a few epochs, while an adaptive band (green) tracking the moving mean stays aligned.}
    \label{fig:scheduling_gap}
\end{figure}

Fixed thresholds implicitly assume that sample difficulty is static, but reward-derived difficulty depends on the current policy and evolves with training. Figure~\ref{fig:scheduling_gap} traces sample difficulty across 14 training epochs on a fixed dataset. As the policy improves, the bulk of the difficulty distribution descends below the static selection band, whereas an adaptive band sliding with the moving mean remains aligned throughout training. Proximal curriculum theory argues that effective scheduling should follow the model's current capability boundary \citep{tzannetos2023proximal}, which requires difficulty estimates that co-evolve with the policy.

Fine-grained rewards typically aggregate component scores additively across tool name, argument key, and argument value. Such additive aggregation ignores the dependency structure of tool execution, since argument-level correctness has no semantic meaning when the tool name is wrong. Figure~\ref{fig:introduction}(b, left) illustrates the resulting leakage, where a wrong tool name can still earn substantial reward when its arguments coincidentally match the reference. A prerequisite-aware reward design avoids the leakage by conditioning argument credit on correct tool selection.

To address both gaps, we propose MATCH (\underline{M}odel-\underline{A}ware \underline{T}ool Learning with \underline{C}urriculum Scheduling and \underline{H}ierarchically Gated Rewards; Figure~\ref{fig:introduction}(c)), a closed-loop framework for RL-based tool learning. Its Model-Aware Curriculum Learning (MACL) component addresses the scheduling gap by maintaining reward-derived difficulty estimates that co-evolve with the policy, selecting samples around the current capability boundary while also drawing harder cases to expand it outward. Its Hierarchical Tool-call Gated Reward (HTGR) component addresses the reward gap by scoring tool-call components hierarchically across tool name, argument key, and argument value, granting credit at each level only when all earlier levels are correct, providing denser feedback than binary rewards while avoiding the spurious credit produced by additive aggregation. The two components form a closed loop, where HTGR produces reward signals used to update the policy and MACL converts the same signals into difficulty estimates that drive sample selection in the next iteration.

Experiments on API-Bank and BFCL V3 show that MATCH outperforms supervised fine-tuning, the binary Tool-N1 baseline, ToolRL, and ToolSample. Ablations confirm complementary gains from MACL and HTGR. The same findings hold across four backbones spanning two families.

Our main contributions are summarized as follows:
\begin{itemize}
    \item We identify two gaps in RL-based tool learning, namely a scheduling gap from fixed difficulty thresholds and a reward gap from additive credit aggregation.
    \item We propose Model-Aware Curriculum Learning (MACL), which schedules samples around the policy's current capability boundary using reward-derived difficulty estimates.
    \item We propose a Hierarchical Tool-call Gated Reward (HTGR), which grants credit across tool name, argument key, and argument value only when all earlier levels are correct.
    \item We empirically validate the effectiveness and generalization of MATCH through extensive experiments and ablations.
\end{itemize}

\section{Related Work}

\subsection{Tool Learning with Reinforcement Learning}

Early tool learning methods rely on prompting or supervised fine-tuning over tool-use trajectories \citep{yao2022react,schick2023toolformer,patil2023gorillalargelanguagemodel,qin2024toolllm,zhang2024xlam}. Such static imitation generalizes poorly to unseen tools and unfamiliar argument structures \citep{zhang2025tooln1,zeng2025toolzero}. Reinforcement learning instead optimizes the policy from tool-execution feedback, and has been applied to search-based reasoning, tool-integrated mathematical reasoning, deep research agents, and multi-turn interaction \citep{jin2025searchr1trainingllmsreason,li2025torlscalingtoolintegratedrl,zheng-etal-2025-deepresearcher,ding2025empoweringmultiturntoolintegratedreasoning}.

The effectiveness of RL hinges on the reward signal, and recent tool-learning rewards diverge along three axes. The first axis is the granularity of correctness decomposition. Tool-N1 keeps the reward binary at the outcome level \citep{zhang2025tooln1}, ToolRL decomposes correctness additively along tool-name, argument-key, and argument-value dimensions \citep{qian2025toolrlrewardtoollearning}, Tool-Zero adapts the strictness of this decomposition across training stages \citep{zeng2025toolzero}, and StepTool moves the unit of decomposition from components to interaction steps by combining call-success and step-contribution scores \citep{yu2025steptool}. The second axis extends the reward with auxiliary objectives, such as the call-efficiency penalty in OTC-PO \citep{wang2025actingreasoningmoreteaching} and the mixed correctness, format, and tool-misuse signal in AutoTIR \citep{wei2025autoTIR}. The third axis moves shaping out of the reward and into advantage shaping for tool-integrated reasoning \citep{lin2025usTIR:TIR}. Process-level supervision in general reasoning shares the same motivation of denser intermediate signals \citep{lightman2023letsverifystepstep}.

\subsection{Curriculum Learning and Dynamic Sampling for RL}

Curriculum learning organizes training examples by difficulty, following the intuition that models learn most effectively from tasks near their current capability boundary \citep{bengio2009curriculum}. Subsequent work develops this principle through competence-based and proximal curricula \citep{platanios2019competence,tzannetos2023proximal}, drawing on the zone of proximal development \citep{vygotsky1978mind}. Dynamic sampling extends the idea to RL by filtering or prioritizing rollouts for training efficiency and stability. Online difficulty filtering removes overly easy or difficult samples to keep training within the informative regime \citep{bae2025onlinedifficulty}, and DAPO integrates dynamic sampling into policy optimization to address scaling challenges in RL \citep{yu2025dapoopensourcellmreinforcement}.

For tool learning, iTool combines a fixed easy-to-hard warm-up curriculum with iterative MCTS-based preference optimization, treating sample difficulty as a pre-defined attribute fixed before training \citep{zeng2025itool}. ToolSample combines reward mean and variance under manually set thresholds to identify informative samples, but the thresholds stay constant while the policy improves, so the selected band drifts off the informative region \citep{feng2025toolsample}.

\section{Preliminaries}
\paragraph{Task formulation.}
Let $\mathcal{D}={x_i}_{i=1}^{|\mathcal{D}|}$ denote the training set, where each sample $x_i=(q_i,\mathcal{Z}_i,\mathcal{G}_i)$ consists of a user query $q_i$, a set of available tools $\mathcal{Z}_i$, and a reference tool-call sequence $\mathcal{G}_i$. Each tool is defined by its name, natural-language description, and parameter schema. Given $q_i$ and $\mathcal{Z}_i$, the policy samples a response $y_i\sim\pi_\theta(\cdot\mid q_i,\mathcal{Z}_i)$. We then parse the tool-call block in $y_i$ into a predicted tool-call sequence $\mathcal{P}_i=\operatorname{parse}(y_i)$, which is compared with the reference sequence $\mathcal{G}_i$ for evaluation.

\section{Method}

\subsection{Method Overview}

\begin{figure*}[t]
    \centering
    \includegraphics[width=0.98\textwidth]{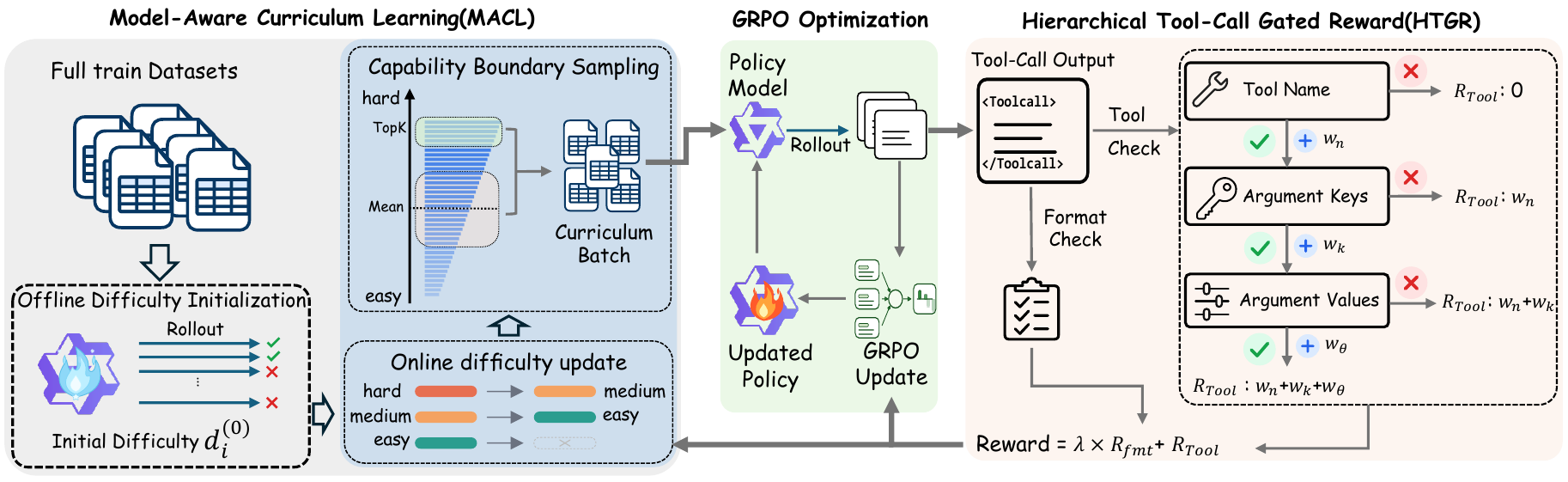}
     \caption{Overview of MATCH. MACL (left) estimates sample difficulty and selects a curriculum batch near the policy's capability boundary, supplemented with harder
  samples. GRPO (middle) updates the policy on the selected batch using HTGR rewards. HTGR (right) evaluates output format and tool-call correctness through hierarchical
  gating, while its rewards refresh sample difficulty for subsequent curriculum selection.}
    \label{fig:method_overview}
\end{figure*}

MATCH jointly addresses the selection of samples for policy optimization and the evaluation of generated tool calls in reinforcement learning for tool use. As illustrated in Figure~\ref{fig:method_overview}, MACL selects training samples according to the policy's evolving capability boundary, while HTGR provides hierarchically gated rewards for the sampled responses. GRPO optimizes the policy using these rewards, which are subsequently fed back to MACL to update sample difficulty, forming a closed training loop.

\subsection{Model-Aware Curriculum Learning}
\label{subsec:macl}

MACL replaces a fixed easy-to-hard schedule with one that tracks the policy's evolving capability boundary. It initializes sample difficulty from offline rollouts of the initial policy, refreshes the estimates online with HTGR reward feedback, and at each epoch selects a boundary band around the current mean difficulty together with a top-$k$ pool of harder cases, which keeps updates near it while expanding it outward.

\paragraph{Offline difficulty initialization.}
For each $x_i$, we run $N$ stochastic rollouts of the initial policy and let $c_i^{(j)}\in\{0,1\}$ indicate whether the $j$-th rollout exactly matches the reference tool-call sequence on both tool identities and argument dictionaries. The rollout accuracy is
\begin{equation}
\alpha_i=\frac{1}{N}\sum_{j=1}^{N}c_i^{(j)}.
\end{equation}
We set the initial smoothed reward $\hat{R}_i^{(-1)}=\alpha_i$ and the initial difficulty $d_i^{(0)}=1-\hat{R}_i^{(-1)}$. We use exact-match success rather than an HTGR-graded score here because the pre-RL policy is not yet aligned with the HTGR signal; the subsequent online updates refine the estimate.

\paragraph{Online difficulty refresh.}
For each $x_i\in\mathcal{S}_t$ at epoch $t$, the $M$ GRPO rollout rewards constitute the group
\begin{equation}
\mathcal{R}_i^{(t)}=\{R_{i,1}^{(t)},\ldots,R_{i,M}^{(t)}\},\quad
\bar{R}_i^{(t)}=\frac{1}{M}\sum_{j=1}^{M}R_{i,j}^{(t)}.
\end{equation}
We normalize the mean to $[0,1]$,
\begin{equation}
\tilde{R}_i^{(t)}=\operatorname{clip}\!\left(\frac{\bar{R}_i^{(t)}-R_{\min}}{R_{\max}-R_{\min}+\varepsilon_{\mathrm{norm}}},0,1\right),
\end{equation}
where $R_{\min}, R_{\max}$ are the HTGR-induced reward bounds and $\varepsilon_{\mathrm{norm}}$ a small constant. An exponential moving average reduces rollout-induced variance in the difficulty estimate,
\begin{equation}
\hat{R}_i^{(t)}=\eta\hat{R}_i^{(t-1)}+(1-\eta)\tilde{R}_i^{(t)},
\end{equation}
and the next-epoch difficulty is $d_i^{(t+1)}=1-\hat{R}_i^{(t)}$. For unselected samples, $\hat{R}_i^{(t)}=\hat{R}_i^{(t-1)}$.

\paragraph{Curriculum sample selection.}
At epoch $t$, the mean difficulty
\begin{equation}
\mu_t=\frac{1}{|\mathcal{D}|}\sum_{i=1}^{|\mathcal{D}|}d_i^{(t)}
\end{equation}
is taken as the current estimate of the capability boundary. The boundary band and above-mean pool are
\begin{equation}
\mathcal{B}_t=\{x_i\in\mathcal{D}:|d_i^{(t)}-\mu_t|\le\epsilon_{\mathrm{band}}\},
\end{equation}
\begin{equation}
\mathcal{C}_t^{+}=\{x_i\in\mathcal{D}\setminus\mathcal{B}_t:d_i^{(t)}>\mu_t\},
\end{equation}
where $\epsilon_{\mathrm{band}}$ sets the band width. The hard subset is selected by top-$k$ sampling over the pool,
\begin{equation}
\mathcal{H}_t=\operatorname{TopK}_{k_t}(\mathcal{C}_t^{+};d_i^{(t)}),\quad
k_t=\left\lfloor r_{\mathrm{hard}}|\mathcal{C}_t^{+}|\right\rfloor,
\end{equation}
where $r_{\mathrm{hard}}\in[0,1]$ is held fixed across epochs. The training subset is $\mathcal{S}_t=\mathcal{B}_t\cup\mathcal{H}_t$.

\subsection{Hierarchical Tool-call Gated Reward}
\label{sec:htgr}

\paragraph{Composite reward and format validity.}
For tool-call samples, HTGR converts each generated response into a scalar reward for both GRPO optimization and MACL difficulty updates. We decompose the reward into an auxiliary format term and a tool-call gating term,
\begin{equation}
R(x,y)=\lambda R_{\mathrm{fmt}}(x,y)+R_{\mathrm{tool}}(x,y),
\end{equation}
where $R_{\mathrm{fmt}}$ checks whether the response follows the required tag structure and $R_{\mathrm{tool}}$ evaluates the correctness of the generated tool calls. Let $\phi_{\mathrm{fmt}}(x,y)$ denote the event that $y$ satisfies the tag constraints required by $x$. The format reward is
\begin{equation}
R_{\mathrm{fmt}}(x,y)=
\begin{cases}
1, & \phi_{\mathrm{fmt}}(x,y),\\
-1, & \neg\phi_{\mathrm{fmt}}(x,y).
\end{cases}
\end{equation}
We use a small fixed $\lambda$ so that format validity contributes only an auxiliary signal while tool-call correctness remains the primary reward. We restrict RL optimization to tool-call samples and exclude direct-response samples from policy updates.

\paragraph{Hierarchical tool-call consistency.}
HTGR enforces tool-name gating. Argument-key and argument-value checks are evaluated only for predictions with the correct tool name. A purely binary reward is sparse during RL optimization, while an ungated additive reward suffers from the credit leakage illustrated in Figure~\ref{fig:introduction}(b), where arguments attached to an incorrect tool still earn partial reward. HTGR therefore uses a dense but tool-name-gated score. After canonical JSON parsing and argument normalization inside the \texttt{<tool\_call>} block, let the reference and predicted tool-call sequences be
\begin{equation}
\mathcal{G}=(g_1,\ldots,g_{m_{\mathrm{ref}}}),\quad \mathcal{P}=(p_1,\ldots,p_{m_{\mathrm{pred}}}).
\end{equation}
For a reference call $g_i$ and a predicted call $p_j$, let $n_i$ and $\hat{n}_j$ denote their tool names, $k_i$ and $\hat{k}_j$ their argument-key sets, and $\theta_i$ and $\hat{\theta}_j$ their argument dictionaries. The pairwise tool-call consistency score is
\begin{equation}
s(g_i,p_j)=
\begin{cases}
w_n+\Delta_k(i,j)+\Delta_{\theta}(i,j), & n_i=\hat{n}_j,\\
0, & n_i\ne\hat{n}_j,
\end{cases}
\end{equation}
where the argument-key increment is
\begin{equation}
\Delta_k(i,j)=
\begin{cases}
w_k, & k_i=\hat{k}_j,\\
0, & k_i\ne\hat{k}_j,
\end{cases}
\end{equation}
and the argument-value increment is
\begin{equation}
\Delta_{\theta}(i,j)=
\begin{cases}
w_{\theta}, & k_i=\hat{k}_j \text{ and } \theta_i=\hat{\theta}_j,\\
0, & \text{otherwise}.
\end{cases}
\end{equation}

For multi-call outputs, we greedily match each reference call $g_i$ in order to the unused same-tool prediction with the highest pairwise score. Let $U_{i-1}$ be the used prediction indices before matching $g_i$, and let $C_i=\{j:j\notin U_{i-1},\ n_i=\hat{n}_j\}$. The selected score is
\begin{equation}
r_i=
\begin{cases}
\max_{j\in C_i}s(g_i,p_j), & C_i\ne\emptyset,\\
0, & C_i=\emptyset.
\end{cases}
\end{equation}
When $C_i\ne\emptyset$, we add the maximizing prediction to the used set. The sequence-level tool-call reward is
\begin{equation}
R_{\mathrm{tool}}(x,y)=
\frac{1}{\max(m_{\mathrm{ref}},1)}
\sum_{i=1}^{m_{\mathrm{ref}}}r_i.
\end{equation}
Unmatched predicted calls receive no positive credit and no additional penalty. If a required tool-call block is missing, we assign a fixed negative penalty to $R_{\mathrm{tool}}$.

\subsection{Policy Update}
At each epoch, MATCH applies standard GRPO~\citep{shao2024deepseekmathpushinglimitsmathematical} to the curriculum subset $\mathcal{S}_t$ selected by MACL, without modifying the original objective. For each prompt $x_i\in\mathcal{S}_t$, we sample $M$ responses from the old policy, score them with HTGR, and compute group-normalized advantages. We update the policy with the clipped GRPO surrogate and apply actor-side KL regularization against a reference policy, keeping the KL term outside the HTGR reward. The same reward group updates MACL's online sample difficulty for the next epoch.

\begin{table*}[t]
\centering
\small
\setlength{\tabcolsep}{2.2pt}
\begin{tabular*}{\textwidth}{@{\extracolsep{\fill}}lcccccccccc@{}}
\toprule
& \multicolumn{4}{c}{\textbf{API-Bank}} & \multicolumn{6}{c}{\textbf{BFCL V3}} \\
\cmidrule(lr){2-5} \cmidrule(lr){6-11}
\textbf{Model} & \textbf{Overall} & \textbf{L1} & \textbf{L2} & \textbf{L3} & \textbf{Overall} & \textbf{NL AST} & \textbf{Live} & \textbf{MT} & \textbf{Rel.} & \textbf{Irrel.} \\
\midrule
ToolACE-8B~\citep{liu2024toolace} & -- & 75.94 & 47.41 & -- & 60.44 & 88.94 & 74.99 & 17.38 & 80.49 & 85.71 \\
GPT-4-turbo~\citep{openai2023gpt4} & -- & 72.43 & 39.26 & -- & 61.89 & 88.80 & 76.23 & 24.88 & 73.17 & 79.76 \\
GPT-4o-mini~\citep{openai2023gpt4} & -- & 74.69 & 45.93 & -- & 60.47 & 83.72 & 70.19 & 27.50 & 80.49 & 71.77 \\
GPT-3.5-turbo~\citep{ouyang2022training} & -- & 70.43 & 52.59 & -- & 53.00 & 78.52 & 61.22 & 19.25 & 97.56 & 35.16 \\
LLaMA-3.1-8B~\citep{dubey2024llama3} & -- & 71.18 & 37.04 & -- & 50.15 & 81.15 & 57.93 & 11.38 & 78.05 & 41.62 \\
\midrule
Qwen2.5-7B (Raw) & 62.48 & 70.68 & 49.25 & \underline{44.27} & 41.97 & 66.02 & 53.51 & 4.25 & 76.47 & 62.66 \\
Qwen2.5-7B (SFT400) & 50.59 & 55.89 & 50.75 & 34.35 & 34.08 & 69.29 & 41.40 & 0.00 & \underline{94.44} & 8.11 \\
Qwen2.5-7B (SFT4k) & 47.07 & 51.13 & 34.33 & 41.22 & 36.53 & 45.15 & 57.13 & 0.75 & 72.22 & 72.32 \\
Qwen2.5-7B (SFT400+GRPO) & 54.10 & 61.40 & 52.24 & 32.82 & 39.25 & 80.69 & 46.51 & 0.25 & \textbf{100.00} & 14.19 \\
Tool-N1~\citep{zhang2025tooln1} & 60.47 & 70.68 & 46.27 & 36.64 & 53.91 & 77.50 & 73.39 & 10.00 & 77.78 & 77.85 \\
ToolRL~\citep{qian2025toolrlrewardtoollearning} & 64.32 & \underline{74.44} & 61.19 & 35.11 & 57.73 & \underline{86.70} & 75.26 & 11.25 & 83.33 & 76.02 \\
ToolSample~\citep{feng2025toolsample} & \underline{64.99} & 73.18 & \textbf{65.67} & 39.69 & \underline{60.25} & 86.42 & \underline{76.85} & \underline{18.50} & 83.33 & \underline{78.44} \\
\midrule
\textbf{MATCH} & \textbf{72.19} & \textbf{76.44} & \textbf{65.67} & \textbf{62.60} & \textbf{62.87} & \textbf{88.61} & \textbf{77.39} & \textbf{22.63} & 83.33 & \textbf{80.39} \\
\bottomrule
\end{tabular*}

\caption{Main results on API-Bank and BFCL V3. All numbers are accuracies (\%). NL AST denotes Non-Live AST, MT denotes Multi Turn, and Rel./Irrel. denote relevance/irrelevance detection.}
\label{tab:main_results}
\end{table*}

\section{Experiments}
We organize the experiments around five research questions.
\textbf{RQ1.} How does MATCH compare with existing tool-learning baselines?
\textbf{RQ2.} How do the components and design choices of MATCH affect its performance?
\textbf{RQ3.} How sensitive is MATCH to key hyperparameters? 
\textbf{RQ4.} Which curriculum sampling strategy works best? 
\textbf{RQ5.} Does MATCH provide consistent gains over raw and RL-trained baselines across different backbone models?

\subsection{Experimental Settings}

\paragraph{Training datasets and benchmarks.}
We use the same 4,000 training set as ToolRL~\citep{qian2025toolrlrewardtoollearning}, consisting of 2,000 ToolACE examples, 1,000 Hammer examples with masked function names, and 1,000 xLAM examples. We evaluate on API-Bank~\citep{li2023apibank} and BFCL V3~\citep{patil2023gorillalargelanguagemodel}, which cover tool-call correctness, live execution, relevance detection, and multi-turn tool use.

\paragraph{Implementation details.}
Unless otherwise stated, we use Qwen2.5-7B-Instruct as the backbone.
We train MATCH with GRPO in veRL~\citep{Sheng2025verl} using 4 NVIDIA A800 GPUs. Following ToolRL, we decompose multi-step trajectories into single-step training instances while retaining prior dialogue history in the prompt. Training uses $M=4$ responses per prompt, 15 epochs, a batch size of 512, and a sampling temperature of 0.7. Actor-side KL regularization uses $\beta_{\mathrm{KL}}=0.001$ with the low-variance KL estimator.


\paragraph{Baselines.}
All main training baselines use Qwen2.5-7B-Instruct and draw their training examples from the same ToolRL dataset. \textbf{Raw} denotes the backbone without additional task-specific training. \textbf{SFT400} and \textbf{SFT4k} use supervised fine-tuning on 400 and 4,000 examples, respectively, while \textbf{SFT400+GRPO} applies GRPO after the 400-example supervised cold start. We further compare with the RL-based baselines \textbf{Tool-N1}~\citep{zhang2025tooln1}, \textbf{ToolRL}~\citep{qian2025toolrlrewardtoollearning}, and \textbf{ToolSample}~\citep{feng2025toolsample}. The additional closed- and open-source models in the upper block of Table~\ref{tab:main_results} are included as reference results only and are excluded from best/second-best annotations.

\begin{table*}[t]
\centering
\small
\setlength{\tabcolsep}{1.8pt}
\begin{tabular*}{\textwidth}{@{\extracolsep{\fill}}lllccccc ccccccc@{}}
\toprule
& & & \multicolumn{5}{c}{\textbf{API-Bank}} & \multicolumn{7}{c}{\textbf{BFCL V3}} \\
\cmidrule(lr){4-8} \cmidrule(lr){9-15}
\textbf{Variant} & \textbf{Sched.} & \textbf{Reward} & \textbf{Ovr.} & \textbf{L1} & \textbf{L2} & \textbf{L3} & \textbf{$\Delta$} & \textbf{Ovr.} & \textbf{NL} & \textbf{Live} & \textbf{MT} & \textbf{Rel.} & \textbf{Irrel.} & \textbf{$\Delta$} \\
\midrule
\textbf{MATCH} & \textbf{MACL} & \textbf{HTGR} & \textbf{72.19} & \textbf{76.44} & \textbf{65.67} & \textbf{62.60} & -- & \textbf{62.87} & \textbf{88.61} & \textbf{77.39} & \textbf{22.63} & \textbf{83.33} & \textbf{80.39} & -- \\
w/o MACL & All data & HTGR & \underline{66.67} & \underline{75.19} & \underline{64.18} & 41.98 & -5.52 & 59.62 & \underline{88.00} & 75.48 & 15.38 & 72.22 & 77.27 & -3.25 \\
w/o HTGR & MACL & Binary & 64.82 & 73.93 & 55.22 & 41.98 & -7.37 & 57.38 & 87.04 & 70.22 & 14.88 & 76.47 & 62.92 & -5.49 \\
w/o HTGR & MACL & ToolRL & 65.33 & 73.18 & 53.73 & \underline{47.33} & -6.86 & \underline{60.05} & 87.57 & \underline{76.32} & \underline{16.25} & 66.67 & \underline{79.50} & -2.82 \\
w/o Both & All data & Binary & 60.47 & 70.68 & 46.27 & 36.64 & -11.72 & 53.91 & 77.50 & 73.39 & 10.00 & \underline{77.78} & 77.85 & -8.96 \\
w/o Both & All data & ToolRL & 64.32 & 74.44 & 61.19 & 35.11 & -7.87 & 57.73 & 86.70 & 75.26 & 11.25 & \textbf{83.33} & 76.02 & -5.14 \\
\bottomrule
\end{tabular*}
\caption{Ablation results on API-Bank and BFCL V3. Accuracy values are reported in percent, and $\Delta$ denotes the percentage-point change in overall accuracy relative to MATCH. Ovr. denotes overall accuracy, NL denotes Non-Live AST, MT denotes Multi Turn, and Rel./Irrel. denote relevance/irrelevance detection.}
\label{tab:ablation_results}
\label{tab:apibank_ablation}
\label{tab:bfcl_ablation}
\end{table*}

\subsection{Main Results (RQ1)}

Table~\ref{tab:main_results} reports results on API-Bank and BFCL V3, comparing MATCH with the raw backbone, supervised fine-tuning variants, and RL-based baselines. The upper block lists additional models for reference. ToolSample is the strongest competing baseline on both benchmarks. MATCH outperforms it by 7.20 percentage points on API-Bank and 2.62 percentage points on BFCL V3 and therefore achieves the highest overall accuracy among the main training baselines.

On API-Bank, MATCH reaches 72.19\% overall accuracy. Relative to ToolSample, it improves L1 by 3.26 percentage points, matches its 65.67\% accuracy on L2, and improves L3 by 22.91 percentage points. The largest gain is therefore observed on L3.

On BFCL V3, MATCH reaches 62.87\% overall accuracy. Compared with ToolSample, it improves Non-Live AST, Live, Multi Turn, and irrelevance detection by 2.19, 0.54, 4.13, and 1.95 percentage points, respectively, while matching its 83.33\% accuracy on relevance detection. Thus, MATCH improves four of the five BFCL V3 categories relative to ToolSample, with the largest category-level gain occurring on Multi Turn.

\subsection{Ablation and Component Analysis (RQ2)}

\paragraph{Main ablation.}
Table~\ref{tab:ablation_results} assesses MACL and HTGR by removing either or both components while keeping the backbone, training set, and all other training settings fixed. Without MACL, training uses the full dataset in every epoch. Without HTGR, MACL remains active, but the HTGR reward is replaced by either a binary correctness reward or ToolRL's additive reward. The joint ablations combine full-data training with each replacement reward.

On API-Bank, removing MACL reduces overall accuracy from 72.19\% to 66.67\%, with the largest decrease occurring on L3. Replacing HTGR with the binary or ToolRL reward reduces overall accuracy to 64.82\% and 65.33\%, respectively. Removing both components further lowers accuracy to 60.47\% with the binary reward and 64.32\% with the ToolRL reward. On BFCL V3, the single-component ablations reduce overall accuracy by 2.82--5.49 percentage points, while the joint ablations yield reductions of 5.14--8.96 percentage points. These results show that both MACL and HTGR contribute to performance across the two benchmarks.

\begin{table}[t]
\centering
\small
\setlength{\tabcolsep}{3.5pt}
\begin{tabular*}{\columnwidth}{@{\extracolsep{\fill}}lcccc@{}}
\toprule
\textbf{Variant} & \textbf{Overall} & \textbf{L1} & \textbf{L2} & \textbf{L3} \\
\midrule
\multicolumn{5}{l}{\textit{Difficulty initialization}} \\
\shortstack[l]{Base-policy rollout\\initialization} & \textbf{72.19} & \textbf{76.44} & \textbf{65.67} & \textbf{62.60} \\
Constant initialization & 68.17 & 75.94 & \textbf{65.67} & 45.80 \\
Random initialization & 65.83 & 74.69 & 58.21 & 42.75 \\
\midrule
\multicolumn{5}{l}{\textit{HTGR reward hierarchy}} \\
Name-only & 64.15 & 73.18 & 62.69 & 37.40 \\
Name + Key & 69.51 & \textbf{77.69} & \textbf{67.16} & 45.80 \\
Full HTGR & \textbf{72.19} & 76.44 & 65.67 & \textbf{62.60} \\
\bottomrule
\end{tabular*}
\caption{Effects of difficulty initialization and the HTGR reward hierarchy on API-Bank. Each block varies only the indicated design choice while keeping the remaining MATCH configuration fixed. All numbers are accuracies (\%).}
\label{tab:design_analysis}
\end{table}

\paragraph{Difficulty initialization.}
On API-Bank, we compare base-policy rollout initialization with constant and random initialization while keeping subsequent online updates and all other settings fixed. The rollout variant estimates $d_i^{(0)}$ from $N=128$ zero-shot rollouts, whereas the alternatives set $d_i^{(0)}=0.5$ or sample it independently from $\mathcal{U}(0,1)$. As shown in Table~\ref{tab:design_analysis}, rollout initialization performs best overall, with its largest advantage on L3. This suggests that policy-aligned initialization provides a more informative starting curriculum and that its influence is not fully eliminated by online updates.

\paragraph{HTGR reward hierarchy.}
On API-Bank, we compare three cumulative HTGR configurations by setting inactive weights to zero and retaining the default active weights without renormalization. Name-only, Name + Key, and Full HTGR use $(w_n,w_k,w_\theta)=(0.4,0,0)$, $(0.4,0.3,0)$, and $(0.4,0.3,0.3)$, respectively. All other settings remain fixed. As shown in Table~\ref{tab:design_analysis}, Name + Key yields the highest L1 and L2 accuracy, whereas Full HTGR performs best overall and on L3. These results support retaining the value-level reward in the full hierarchy, particularly for L3.

\subsection{Hyperparameter Analysis (RQ3)}
\label{sec:hyperparam}

\paragraph{MACL scheduling hyperparameters.}
On API-Bank, we analyze sensitivity to the boundary-band width $\epsilon_{\mathrm{band}}$ and the hard-sample ratio $r_{\mathrm{hard}}$, which respectively control the width of the near-boundary selection band and the fraction of above-mean candidates added as hard samples. All other settings are fixed at their default values.

\begin{figure}[t]
    \centering
    \begin{minipage}[t]{0.43\columnwidth}
        \centering
        \includegraphics[width=\linewidth]{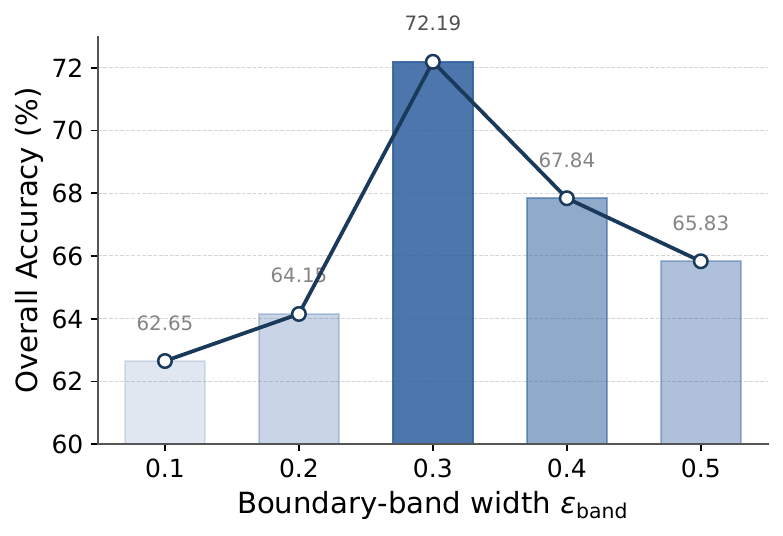}
        \par\smallskip
        \textbf{(a)}
    \end{minipage}\hfill
    \begin{minipage}[t]{0.52\columnwidth}
        \centering
        \includegraphics[width=\linewidth]{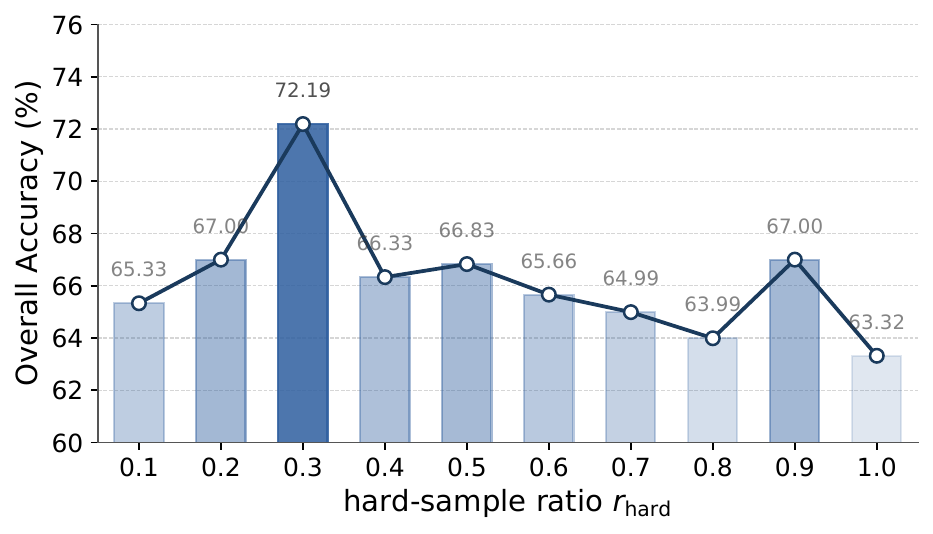}
        \par\smallskip
        \textbf{(b)}
    \end{minipage}
    \caption{Sensitivity of API-Bank overall accuracy to the MACL scheduling hyperparameters. \textbf{(a)} Boundary-band width $\epsilon_{\mathrm{band}}$. \textbf{(b)} Hard-sample ratio $r_{\mathrm{hard}}$.}
    \label{fig:macl_scheduling_sensitivity}
\end{figure}

Figure~\ref{fig:macl_scheduling_sensitivity}(a) shows that overall accuracy peaks at $\epsilon_{\mathrm{band}}=0.3$. Narrower and wider bands both reduce accuracy, suggesting that an overly narrow band limits sample diversity whereas an overly broad band weakens the focus on the current capability boundary. We use $\epsilon_{\mathrm{band}}=0.3$ as the default.

Figure~\ref{fig:macl_scheduling_sensitivity}(b) likewise shows that overall accuracy peaks at $r_{\mathrm{hard}}=0.3$. Performance is lower at both smaller and larger ratios, suggesting that too few hard samples limit exposure to challenging cases whereas excessive hard-sample exposure dilutes near-boundary refinement. We use $r_{\mathrm{hard}}=0.3$ as the default.

\paragraph{HTGR weight allocation.}
On API-Bank, we further examine whether HTGR depends on a particular allocation of the three reward weights. We keep the gating structure, MACL scheduler, training set, and training budget unchanged and vary only $(w_n,w_k,w_\theta)$. All tested allocations retain the complete three-level reward hierarchy and sum to one.

\begin{table}[t]
\centering
\small
\setlength{\tabcolsep}{3.5pt}
\begin{tabular*}{\columnwidth}{@{\extracolsep{\fill}}lcccc@{}}
\toprule
\textbf{Weight Setting} & $\boldsymbol{w_n}$ & $\boldsymbol{w_k}$ & $\boldsymbol{w_\theta}$ & \textbf{Overall} \\
\midrule
Uniform & 0.34 & 0.33 & 0.33 & 69.49 \\
Name-heavy & 0.60 & 0.20 & 0.20 & 71.50 \\
Moderate name-heavy & 0.50 & 0.25 & 0.25 & \textbf{72.50} \\
Key-heavy & 0.20 & 0.60 & 0.20 & 69.99 \\
Key--value-heavy & 0.20 & 0.40 & 0.40 & 70.66 \\
Value-heavy & 0.20 & 0.20 & 0.60 & 69.32 \\
Default HTGR & 0.40 & 0.30 & 0.30 & \underline{72.19} \\
\bottomrule
\end{tabular*}
\caption{Sensitivity of API-Bank overall accuracy to the HTGR weight allocation. The default setting is used in all main experiments.}
\label{tab:htgr_weight_sensitivity}
\end{table}

As shown in Table~\ref{tab:htgr_weight_sensitivity}, overall accuracy ranges from 69.32\% to 72.50\% across the tested allocations. The Moderate name-heavy setting achieves the highest accuracy among the tested settings, only 0.31 percentage points above the prespecified default setting of $(0.4,0.3,0.3)$. The default setting was fixed before this sensitivity analysis and is used in all main experiments; the analysis is not used for post-hoc configuration selection. Moreover, all complete HTGR variants outperform the MACL variant using the ToolRL reward, which achieves 65.33\% in Table~\ref{tab:ablation_results}. Taken together, these results show that complete HTGR retains an advantage across weight allocations and that the prespecified default is close to the best tested setting.

\subsection{Curriculum Sampling Strategy Analysis (RQ4)}

We evaluate whether the MACL sampling strategy is preferable to simpler sampling alternatives. MACL combines boundary-band sampling with a limited top-$k$ hard-sample subset. We compare it with three variants: boundary-band sampling without the top-$k$ subset, selecting all below-mean samples, and selecting all above-mean samples. All variants use the same difficulty updates and training settings and differ only in the sampling rule.

\begin{figure}[t]
    \centering
    \includegraphics[width=\columnwidth]{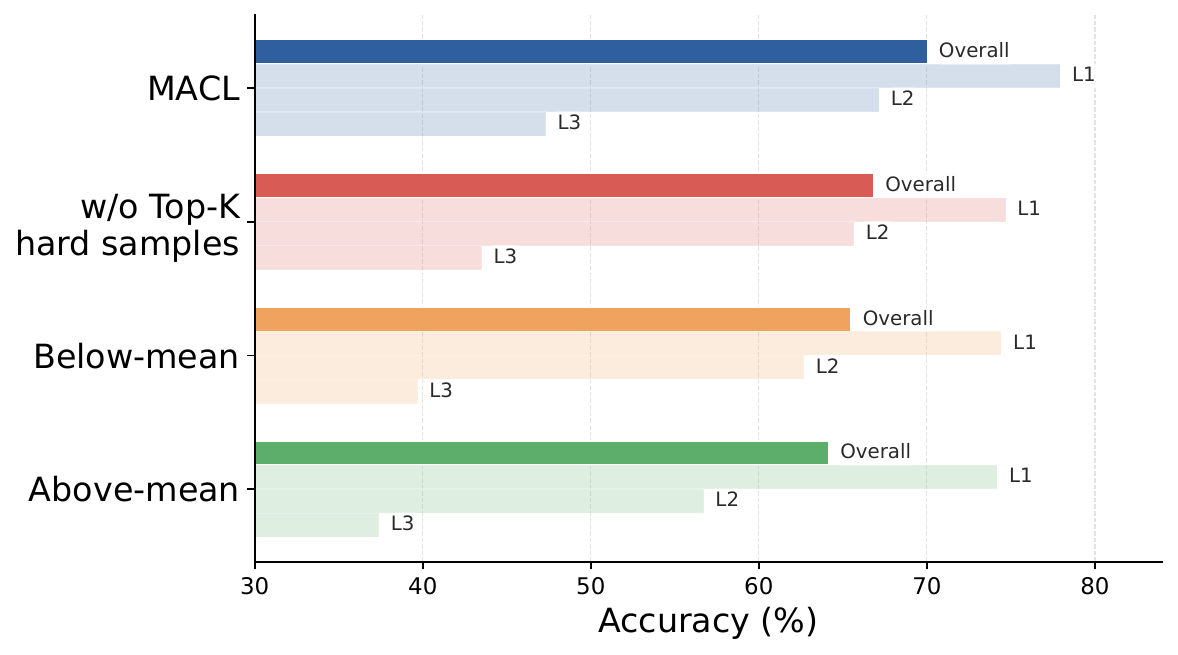}
    \caption{Accuracy of different curriculum sampling strategies on API-Bank. MACL denotes the proposed sampling strategy.}
    \label{fig:curriculum_sampling_strategy}
\end{figure}

Figure~\ref{fig:curriculum_sampling_strategy} shows that MACL achieves the best overall performance, outperforming the alternatives by 5.36--8.04 percentage points. Its largest gain occurs on L3, where removing the top-$k$ hard-sample subset reduces accuracy from 62.60\% to 43.51\%. The weaker below-mean and above-mean variants further suggest that combining near-boundary samples with limited hard-sample exposure is more effective than selecting only one side of the capability boundary.

\subsection{Backbone Generalization Analysis (RQ5)}

We further examine whether MATCH improves tool calling relative to the raw backbone and prior RL training baselines. Following the backbone-wise API-Bank comparison in ToolRL~\citep{qian2025toolrlrewardtoollearning}, we report Raw, SFT400, SFT4k, SFT400+GRPO, and ToolRL results for each backbone, where the baseline numbers follow the values reported by~\citet{qian2025toolrlrewardtoollearning} under the ``Ours, GRPO Cold Start'' setting. The Qwen2.5-7B-Instruct ToolRL row therefore differs from our in-house reproduction in Table~\ref{tab:main_results}. Except for the backbone, all MATCH runs use the same training set and training settings. This comparison evaluates whether the proposed reward-scheduling design remains effective across different model scales and model families.

\begin{table}[t]
\centering
\small
\setlength{\tabcolsep}{2pt}
\begin{tabular}{llcccc}
\toprule
\textbf{Backbone} & \textbf{Setting} & \textbf{Overall} & \textbf{L1} & \textbf{L2} & \textbf{L3} \\
\midrule
\multirow{6}{*}{\shortstack[l]{Qwen2.5-1.5B-\\Instruct}} & Raw & 30.65 & 28.32 & 35.82 & 35.11 \\
 & SFT400 & 53.60 & 57.14 & 50.75 & 44.27 \\
 & SFT4k & 47.07 & 52.88 & 52.24 & 26.72 \\
 & SFT400+GRPO & 61.31 & 64.16 & 58.21 & \textbf{54.20} \\
 & ToolRL & 63.15 & 70.68 & \textbf{61.19} & 41.22 \\
 & \textbf{MATCH} & \textbf{63.82} & \textbf{71.68} & 59.70 & 41.98 \\
\midrule
\multirow{6}{*}{\shortstack[l]{Qwen2.5-3B-\\Instruct}} & Raw & 51.59 & 59.65 & 32.84 & 36.64 \\
 & SFT400 & 52.76 & 59.65 & 50.75 & 32.82 \\
 & SFT4k & 50.92 & 55.64 & 43.28 & 40.46 \\
 & SFT400+GRPO & 62.48 & 68.67 & 58.21 & 45.80 \\
 & ToolRL & 67.00 & \textbf{73.43} & \textbf{67.16} & 47.33 \\
 & \textbf{MATCH} & \textbf{68.17} & \textbf{73.43} & 61.19 & \textbf{55.73} \\
\midrule
\multirow{6}{*}{\shortstack[l]{Qwen2.5-7B-\\Instruct}} & Raw & 62.48 & 70.68 & 49.25 & 44.27 \\
 & SFT400 & 50.59 & 55.89 & 50.75 & 34.35 \\
 & SFT4k & 47.07 & 51.13 & 34.33 & 41.22 \\
 & SFT400+GRPO & 54.10 & 61.40 & 52.24 & 32.82 \\
 & ToolRL & 64.66 & 73.93 & 61.19 & 38.17 \\
 & \textbf{MATCH} & \textbf{72.19} & \textbf{76.44} & \textbf{65.67} & \textbf{62.60} \\
\midrule
\multirow{6}{*}{\shortstack[l]{Llama-3.2-3B-\\Instruct}} & Raw & 40.54 & 44.86 & 29.85 & 32.82 \\
 & SFT400 & 52.76 & 60.65 & 35.82 & 37.40 \\
 & SFT4k & 43.89 & 53.88 & 29.85 & 20.61 \\
 & SFT400+GRPO & 56.78 & 63.60 & 41.79 & 43.51 \\
 & ToolRL & 59.13 & 65.66 & 52.24 & 42.75 \\
 & \textbf{MATCH} & \textbf{65.66} & \textbf{70.18} & \textbf{62.69} & \textbf{53.44} \\
\bottomrule
\end{tabular}
\caption{API-Bank results on different backbone models. MATCH labels and the best metric values within each backbone are bold.}
\label{tab:backbone_generalization}
\end{table}

Table~\ref{tab:backbone_generalization} shows that MATCH achieves higher mean overall accuracy than ToolRL across all four backbones, including Qwen2.5-1.5B-Instruct, Qwen2.5-3B-Instruct, Qwen2.5-7B-Instruct, and Llama-3.2-3B-Instruct. 
These results indicate that the effectiveness of MATCH is not restricted to a particular model scale or backbone family.

\section{Conclusion}


We presented MATCH, a closed-loop framework for RL-based tool learning that couples model-aware curriculum scheduling with hierarchically gated rewards. MACL maintains reward-derived, policy-dependent difficulty estimates and selects samples near the evolving capability boundary together with a limited top-$k$ hard-sample subset. HTGR scores tool name, argument key, and argument value through prerequisite gates, preventing argument-level credit when tool selection is incorrect. On API-Bank and BFCL V3, MATCH achieves 72.19\% and 62.87\% overall accuracy, outperforming all main training baselines, with its largest gains on API-Bank L3. Ablations show that MACL and HTGR each contribute to performance, while controlled analyses support rollout-based difficulty initialization and the complete three-level reward hierarchy. Backbone experiments further show that MATCH achieves higher mean overall accuracy than ToolRL across the tested model scales and families.

\bibliography{custom}

@misc{lin2025usTIR:TIR,
      title={Understanding Tool-Integrated Reasoning},
      author={Heng Lin and Zhongwen Xu},
      year={2025},
      eprint={2508.19201},
      archivePrefix={arXiv},
      primaryClass={cs.LG},
      url={https://arxiv.org/abs/2508.19201},
}

@inproceedings{li2023apibank,
  author    = {Li, M. and Zhao, Y. and Yu, B. and Song, F. and Li, H. and Yu, H. and Li, Z. and Huang, F. and Li, Y.},
  title     = {API-Bank: A Comprehensive Benchmark for Tool-Augmented LLMs},
  booktitle = {Proceedings of the 2023 Conference on Empirical Methods in Natural Language Processing},
  year      = {2023}
}

@misc{qian2025toolrlrewardtoollearning,
      title={ToolRL: Reward is All Tool Learning Needs},
      author={Cheng Qian and Emre Can Acikgoz and Qi He and Hongru Wang and Xiusi Chen and Dilek Hakkani-Tür and Gokhan Tur and Heng Ji},
      year={2025},
      eprint={2504.13958},
      archivePrefix={arXiv},
      primaryClass={cs.LG},
      url={https://arxiv.org/abs/2504.13958},
}

@misc{jin2025searchr1trainingllmsreason,
      title={Search-R1: Training LLMs to Reason and Leverage Search Engines with Reinforcement Learning},
      author={Bowen Jin and Hansi Zeng and Zhenrui Yue and Jinsung Yoon and Sercan Arik and Dong Wang and Hamed Zamani and Jiawei Han},
      year={2025},
      eprint={2503.09516},
      archivePrefix={arXiv},
      primaryClass={cs.CL},
      url={https://arxiv.org/abs/2503.09516},
}

@misc{wang2025actingreasoningmoreteaching,
      title={Acting Less is Reasoning More! Teaching Model to Act Efficiently},
      author={Hongru Wang and Cheng Qian and Wanjun Zhong and Xiusi Chen and Jiahao Qiu and Shijue Huang and Bowen Jin and Mengdi Wang and Kam-Fai Wong and Heng Ji},
      year={2025},
      eprint={2504.14870},
      archivePrefix={arXiv},
      primaryClass={cs.AI},
      url={https://arxiv.org/abs/2504.14870},
}

@misc{li2025torlscalingtoolintegratedrl,
      title={ToRL: Scaling Tool-Integrated RL},
      author={Xuefeng Li and Haoyang Zou and Pengfei Liu},
      year={2025},
      eprint={2503.23383},
      archivePrefix={arXiv},
      primaryClass={cs.CL},
      url={https://arxiv.org/abs/2503.23383},
}

@misc{ding2025empoweringmultiturntoolintegratedreasoning,
      title={Empowering Multi-Turn Tool-Integrated Reasoning with Group Turn Policy Optimization},
      author={Yifeng Ding and Hung Le and Songyang Han and Kangrui Ruan and Zhenghui Jin and Varun Kumar and Zijian Wang and Anoop Deoras},
      year={2025},
      eprint={2511.14846},
      archivePrefix={arXiv},
      primaryClass={cs.LG},
      url={https://arxiv.org/abs/2511.14846},
}

@inproceedings{yao2022react,
  author    = {Yao, S. and Zhao, J. and Yu, D. and Du, N. and Shafran, I. R. and Narasimhan, K. R. and Cao, Y.},
  title     = {{ReAct}: Synergizing Reasoning and Acting in Language Models},
  booktitle = {Proceedings of the Eleventh International Conference on Learning Representations},
  year      = {2023}
}

@misc{chen2023afireact,
      title={FireAct: Toward Language Agent Fine-tuning},
      author={Baian Chen and Chang Shu and Ehsan Shareghi and Nigel Collier and Karthik Narasimhan and Shunyu Yao},
      year={2023},
      eprint={2310.05915},
      archivePrefix={arXiv},
      primaryClass={cs.CL},
      url={https://arxiv.org/abs/2310.05915},
}

@article{schick2023toolformer,
  author  = {Schick, T. and Dwivedi-Yu, J. and Dess{\`i}, R. and Raileanu, R. and Lomeli, M. and Hambro, E. and Zettlemoyer, L. and Cancedda, N. and Scialom, T.},
  title   = {Toolformer: Language Models Can Teach Themselves to Use Tools},
  journal = {Advances in Neural Information Processing Systems},
  volume  = {36},
  pages   = {68539--68551},
  year    = {2023}
}

@misc{lightman2023letsverifystepstep,
      title={Let's Verify Step by Step},
      author={Hunter Lightman and Vineet Kosaraju and Yura Burda and Harri Edwards and Bowen Baker and Teddy Lee and Jan Leike and John Schulman and Ilya Sutskever and Karl Cobbe},
      year={2023},
      eprint={2305.20050},
      archivePrefix={arXiv},
      primaryClass={cs.LG},
      url={https://arxiv.org/abs/2305.20050},
}

@inproceedings{ouyang2022training,
  title     = {Training Language Models to Follow Instructions with Human Feedback},
  author    = {Ouyang, Long and Wu, Jeffrey and Jiang, Xu and Almeida, Diogo and Wainwright, Carroll and Mishkin, Pamela and Zhang, Chong and Agarwal, Sandhini and Slama, Katarina and Ray, Alex and others},
  booktitle = {Advances in Neural Information Processing Systems},
  volume    = {35},
  pages     = {27730--27744},
  year      = {2022}
}

@misc{shao2024deepseekmathpushinglimitsmathematical,
      title={DeepSeekMath: Pushing the Limits of Mathematical Reasoning in Open Language Models},
      author={Zhihong Shao and Peiyi Wang and Qihao Zhu and Runxin Xu and Junxiao Song and Xiao Bi and Haowei Zhang and Mingchuan Zhang and Y. K. Li and Y. Wu and Daya Guo},
      year={2024},
      eprint={2402.03300},
      archivePrefix={arXiv},
      primaryClass={cs.CL},
      url={https://arxiv.org/abs/2402.03300},
}

@misc{yu2025dapoopensourcellmreinforcement,
      title={DAPO: An Open-Source LLM Reinforcement Learning System at Scale},
      author={Qiying Yu and Zheng Zhang and Ruofei Zhu and Yufeng Yuan and Xiaochen Zuo and Yu Yue and Weinan Dai and Tiantian Fan and Gaohong Liu and Lingjun Liu and Xin Liu and Haibin Lin and Zhiqi Lin and Bole Ma and Guangming Sheng and Yuxuan Tong and Chi Zhang and Mofan Zhang and Wang Zhang and Hang Zhu and Jinhua Zhu and Jiaze Chen and Jiangjie Chen and Chengyi Wang and Hongli Yu and Yuxuan Song and Xiangpeng Wei and Hao Zhou and Jingjing Liu and Wei-Ying Ma and Ya-Qin Zhang and Lin Yan and Mu Qiao and Yonghui Wu and Mingxuan Wang},
      year={2025},
      eprint={2503.14476},
      archivePrefix={arXiv},
      primaryClass={cs.LG},
      url={https://arxiv.org/abs/2503.14476},
}

@misc{wei2025autoTIR,
      title={AutoTIR: Autonomous Tools Integrated Reasoning via Reinforcement Learning},
      author={Yifan Wei and Xiaoyan Yu and Yixuan Weng and Tengfei Pan and Angsheng Li and Li Du},
      year={2025},
      eprint={2507.21836},
      archivePrefix={arXiv},
      primaryClass={cs.CL},
      url={https://arxiv.org/abs/2507.21836},
}

@misc{zhang2025tooln1,
      title={Nemotron-Research-Tool-N1: Exploring Tool-Using Language Models with Reinforced Reasoning},
      author={Shaokun Zhang and Yi Dong and Jieyu Zhang and Jan Kautz and Bryan Catanzaro and Andrew Tao and Qingyun Wu and Zhiding Yu and Guilin Liu},
      year={2025},
      eprint={2505.00024},
      archivePrefix={arXiv},
      primaryClass={cs.CL},
      url={https://arxiv.org/abs/2505.00024},
}

@inproceedings{zheng-etal-2025-deepresearcher,
    title = "{D}eep{R}esearcher: Scaling Deep Research via Reinforcement Learning in Real-world Environments",
    author = "Zheng, Yuxiang  and
      Fu, Dayuan  and
      Hu, Xiangkun  and
      Cai, Xiaojie  and
      Ye, Lyumanshan  and
      Lu, Pengrui  and
      Liu, Pengfei",
    editor = "Christodoulopoulos, Christos  and
      Chakraborty, Tanmoy  and
      Rose, Carolyn  and
      Peng, Violet",
    booktitle = "Proceedings of the 2025 Conference on Empirical Methods in Natural Language Processing",
    month = nov,
    year = "2025",
    address = "Suzhou, China",
    publisher = "Association for Computational Linguistics",
    url = "https://aclanthology.org/2025.emnlp-main.22/",
    doi = "10.18653/v1/2025.emnlp-main.22",
    pages = "414--431",
    ISBN = "979-8-89176-332-6",
}

@misc{qin2024toollearningfoundationmodels,
      title={Tool Learning with Foundation Models},
      author={Yujia Qin and Shengding Hu and Yankai Lin and Weize Chen and Ning Ding and Ganqu Cui and Zheni Zeng and Yufei Huang and Chaojun Xiao and Chi Han and Yi Ren Fung and Yusheng Su and Huadong Wang and Cheng Qian and Runchu Tian and Kunlun Zhu and Shihao Liang and Xingyu Shen and Bokai Xu and Zhen Zhang and Yining Ye and Bowen Li and Ziwei Tang and Jing Yi and Yuzhang Zhu and Zhenning Dai and Lan Yan and Xin Cong and Yaxi Lu and Weilin Zhao and Yuxiang Huang and Junxi Yan and Xu Han and Xian Sun and Dahai Li and Jason Phang and Cheng Yang and Tongshuang Wu and Heng Ji and Zhiyuan Liu and Maosong Sun},
      year={2024},
      eprint={2304.08354},
      archivePrefix={arXiv},
      primaryClass={cs.CL},
      url={https://arxiv.org/abs/2304.08354},
}

@misc{patil2023gorillalargelanguagemodel,
      title={Gorilla: Large Language Model Connected with Massive APIs},
      author={Shishir G. Patil and Tianjun Zhang and Xin Wang and Joseph E. Gonzalez},
      year={2023},
      eprint={2305.15334},
      archivePrefix={arXiv},
      primaryClass={cs.CL},
      url={https://arxiv.org/abs/2305.15334},
}

@inproceedings{qin2024toolllm,
      title={ToolLLM: Facilitating Large Language Models to Master 16000+ Real-world APIs},
      author={Yujia Qin and Shihao Liang and Yining Ye and Kunlun Zhu and Lan Yan and Yaxi Lu and Yankai Lin and Xin Cong and Xiangru Tang and Bill Qian and Sihan Zhao and Lauren Hong and Runchu Tian and Ruobing Xie and Jie Zhou and Mark Gerstein and Dahai Li and Zhiyuan Liu and Maosong Sun},
      booktitle={Proceedings of the Twelfth International Conference on Learning Representations},
      year={2024},
      url={https://openreview.net/forum?id=dHng2O0Jjr},
}

@misc{zhang2024xlam,
      title={xLAM: A Family of Large Action Models to Empower AI Agent Systems},
      author={Jianguo Zhang and Tian Lan and Ming Zhu and Zuxin Liu and Thai Hoang and Shirley Kokane and Weiran Yao and Juntao Tan and Akshara Prabhakar and Haolin Chen and Zhiwei Liu and Yihao Feng and Tulika Awalgaonkar and Rithesh Murthy and Eric Hu and Zeyuan Chen and Ran Xu and Juan Carlos Niebles and Shelby Heinecke and Huan Wang and Silvio Savarese and Caiming Xiong},
      year={2024},
      eprint={2409.03215},
      archivePrefix={arXiv},
      primaryClass={cs.CL},
      url={https://arxiv.org/abs/2409.03215},
}

@inproceedings{Sheng2025verl,
   title={HybridFlow: A Flexible and Efficient RLHF Framework},
   url={http://dx.doi.org/10.1145/3689031.3696075},
   DOI={10.1145/3689031.3696075},
   booktitle={Proceedings of the Twentieth European Conference on Computer Systems},
   publisher={ACM},
   author={Sheng, Guangming and Zhang, Chi and Ye, Zilingfeng and Wu, Xibin and Zhang, Wang and Zhang, Ru and Peng, Yanghua and Lin, Haibin and Wu, Chuan},
   year={2025},
   month=mar, pages={1279--1297},
   collection={EuroSys '25}
}

@article{feng2025toolsample,
  title={ToolSample: Dual Dynamic Sampling Methods with Curriculum Learning for RL-based Tool Learning},
  author={Feng, Zihao and Wang, Xiaoxue and Wu, Bowen and Cao, Hailong and Zhao, Tiejun and Yu, Qun and Wang, Baoxun},
  journal={arXiv preprint arXiv:2509.14718},
  year={2025}
}

@article{bae2025onlinedifficulty,
  title={Online difficulty filtering for reasoning oriented reinforcement learning},
  author={Bae, Sanghwan and Hong, Jiwoo and Lee, Min Young and Kim, Hanbyul and Nam, Jeongyeon and Kwak, Donghyun},
  journal={arXiv preprint arXiv:2504.03380},
  year={2025}
}

@book{vygotsky1978mind,
  title={Mind in society: The development of higher psychological processes},
  author={Vygotsky, Lev S.},
  year={1978},
  publisher={Harvard University Press}
}

@inproceedings{bengio2009curriculum,
  title     = {Curriculum Learning},
  author    = {Bengio, Yoshua and Louradour, J{\'e}r{\^o}me and Collobert, Ronan and Weston, Jason},
  booktitle = {Proceedings of the 26th International Conference on Machine Learning},
  pages     = {41--48},
  year      = {2009},
  publisher = {ACM},
  doi       = {10.1145/1553374.1553380}
}

@inproceedings{platanios2019competence,
  title     = {Competence-based Curriculum Learning for Neural Machine Translation},
  author    = {Platanios, Emmanouil Antonios and Stretcu, Otilia and Neubig, Graham and Poczos, Barnabas and Mitchell, Tom},
  booktitle = {Proceedings of the 2019 Conference of the North American Chapter of the Association for Computational Linguistics: Human Language Technologies, Volume 1 (Long and Short Papers)},
  pages     = {1162--1172},
  year      = {2019},
  address   = {Minneapolis, Minnesota},
  publisher = {Association for Computational Linguistics},
  doi       = {10.18653/v1/N19-1119},
  url       = {https://aclanthology.org/N19-1119/}
}

@article{tzannetos2023proximal,
  title   = {Proximal Curriculum for Reinforcement Learning Agents},
  author  = {Tzannetos, Georgios and Gomes Ribeiro, B{\'a}rbara and Kamalaruban, Parameswaran and Singla, Adish},
  journal = {Transactions on Machine Learning Research},
  year    = {2023},
  url     = {https://openreview.net/forum?id=8WUyeeMxMH}
}

@article{zeng2025toolzero,
  title         = {Tool Zero: Training Tool-Augmented {LLM}s via Pure {RL} from Scratch},
  author        = {Zeng, Yirong and Ding, Xiao and Hou, Yutai and Wang, Yuxian and Du, Li and Dai, Juyi and Ding, Qiuyang and Tang, Duyu and Tu, Dandan and Liu, Weiwen and Qin, Bing and Liu, Ting},
  journal       = {arXiv preprint arXiv:2511.01934},
  year          = {2025},
  eprint        = {2511.01934},
  archivePrefix = {arXiv},
  primaryClass  = {cs.LG}
}

@misc{liu2024toolace,
  title         = {ToolACE: Winning the Points of {LLM} Function Calling},
  author        = {Liu, Weiwen and Huang, Xu and Zeng, Xingshan and Hao, Xinlong and Yu, Shuai and Li, Dexun and Wang, Shuai and Gan, Weinan and Liu, Zhengying and Yu, Yuanqing and Wang, Zezhong and Wang, Yuxian and Ning, Wu and Hou, Yutai and Wang, Bin and Wu, Chuhan and Wang, Xinzhi and Liu, Yong and Wang, Yasheng and Tang, Duyu and Tu, Dandan and Shang, Lifeng and Jiang, Xin and Tang, Ruiming and Lian, Defu and Liu, Qun and Chen, Enhong},
  year          = {2024},
  eprint        = {2409.00920},
  archivePrefix = {arXiv},
  primaryClass  = {cs.CL}
}

@misc{dubey2024llama3,
  title         = {The Llama 3 Herd of Models},
  author        = {{Llama Team}},
  year          = {2024},
  eprint        = {2407.21783},
  archivePrefix = {arXiv},
  primaryClass  = {cs.AI}
}

@misc{openai2023gpt4,
  title         = {{GPT-4} Technical Report},
  author        = {{OpenAI}},
  year          = {2023},
  eprint        = {2303.08774},
  archivePrefix = {arXiv},
  primaryClass  = {cs.CL}
}

@inproceedings{yu2025steptool,
  author    = {Yu, Yuanqing and Wang, Zhefan and Ma, Weizhi and Wang, Shuai and Wu, Chuhan and Guo, Zhiqiang and Zhang, Min},
  title     = {StepTool: Enhancing Multi-Step Tool Usage in {LLM}s via Step-Grained Reinforcement Learning},
  year      = {2025},
  isbn      = {9798400720406},
  publisher = {Association for Computing Machinery},
  address   = {New York, NY, USA},
  url       = {https://doi.org/10.1145/3746252.3761391},
  doi       = {10.1145/3746252.3761391},
  booktitle = {Proceedings of the 34th ACM International Conference on Information and Knowledge Management},
  pages     = {3952--3962},
  numpages  = {11},
  location  = {Seoul, Republic of Korea},
  series    = {CIKM '25}
}

@misc{zeng2025itool,
  title         = {iTool: Reinforced Fine-Tuning with Dynamic Deficiency Calibration for Advanced Tool Use},
  author        = {Yirong Zeng and Xiao Ding and Yuxian Wang and Weiwen Liu and Wu Ning and Yutai Hou and Xu Huang and Duyu Tang and Dandan Tu and Bing Qin and Ting Liu},
  year          = {2025},
  eprint        = {2501.09766},
  archivePrefix = {arXiv},
  primaryClass  = {cs.CL},
  url           = {https://arxiv.org/abs/2501.09766}
}

\end{document}